\documentclass[lettersize,journal]{IEEEtran}
\usepackage{amsmath,amsfonts}
\usepackage{algorithmic}
\usepackage{algorithm}
\usepackage{array}
\usepackage[caption=false,font=normalsize,labelfont=sf,textfont=sf]{subfig}
\usepackage{textcomp}
\usepackage{stfloats}
\usepackage{url}
\usepackage{verbatim}
\usepackage{graphicx}
\usepackage[nocompress]{cite}
\usepackage{multirow}
\usepackage[table]{xcolor}
\usepackage{makecell}
\usepackage{amsmath}
\usepackage{amsthm}

\definecolor{mygray}{RGB}{240,240,240}      %
\definecolor{myblue}{RGB}{230,240,255}
\definecolor{mygreen}{RGB}{240,255,240}
\usepackage{hyperref}
\hypersetup{
    hidelinks=false, %
    colorlinks=true, %
    linkcolor=blue,  %
    citecolor=blue,  %
    urlcolor=blue,   %
    breaklinks=true, %
    bookmarks=true   %
}
\usepackage{booktabs}
\definecolor{myblue}{RGB}{230,242,255}
\begin{document}

\title{Ground4D: Consistency-Aware 4D Reconstruction\\from Monocular Video}

\author{Qing Zhao,~Weijian Deng,~Pengxu Wei,~Liang Lin,~\IEEEmembership{Fellow,~IEEE}
\thanks{Qing Zhao, Pengxu Wei, and Liang Lin are with the School of Computer Science and Engineering, Sun Yat-sen University, China. (e-mail: zhaoq78@mail2.sysu.edu.cn; weipx3@mail.sysu.edu.cn; linliang@ieee.org)}
\thanks{Weijian Deng is with the Tsinghua Shenzhen International Graduate School, Tsinghua University, China. (e-mail: dengwj16@gmail.com)}
}

\markboth{Journal of \LaTeX\ Class Files,~Vol.~14, No.~8, August~2021}%
{Shell \MakeLowercase{\textit{et al.}}: A Sample Article Using IEEEtran.cls for IEEE Journals}

\maketitle

\begin{abstract}
Learning a 4D scene representation from a single monocular video that supports dynamic novel-view synthesis while maintaining faithful geometry over time remains challenging.
Dynamic Gaussian Splatting achieves strong rendering performance through photometric optimization, yet does not explicitly enforce multi-view geometric consistency. In contrast, 3D foundation models recover coherent scene geometry and camera motion, but their point-based outputs are not designed for photorealistic rendering.
We propose Ground4D, a geometry-grounded framework built on two stages. First, we perform geometry initialization via 3D foundation models, leveraging VGGT in a training-free manner to reconstruct multi-view-consistent 3D geometry and camera poses from monocular video. The recovered geometry provides a structured and reliable initialization for dynamic Gaussian representations. Second, we conduct geometry-consistency-aware refinement via dynamic Gaussian Splatting, optimizing the representation through differentiable rendering while maintaining multi-view geometric consistency across both observed and synthesized viewpoints.
Furthermore, Ground4D inherently models the continuous 4D dynamics of the scene, naturally supporting rendering at arbitrary timestamps.
By integrating foundation-level geometric priors into dynamic Gaussian optimization, Ground4D achieves stronger reconstruction fidelity and rendering performance, underscoring the role of geometry-grounded constraints in robust 4D scene modeling.
\end{abstract}


\section{Introduction}
\IEEEPARstart{R}{econstructing} dynamic 4D scenes from monocular video has become an important problem in computer vision, with applications in immersive telepresence, robotics, and content creation. A practical 4D representation should preserve temporally consistent geometry while enabling high-quality novel-view synthesis. Achieving both objectives simultaneously, however, remains challenging.

Recent advances in dynamic Gaussian Splatting have demonstrated impressive rendering performance for dynamic novel-view synthesis~\cite{wu20244d,yang2024deformable,luiten2024dynamic,stearns2024dynamic,lei2025mosca,wang2025shape}. By directly optimizing radiance primitives with photometric supervision, these methods produce visually compelling results. Nevertheless, because geometry is implicitly inferred through image reconstruction losses, purely photometric optimization may lead to structural drift, degraded depth consistency, and instability under large viewpoint variations.

In parallel, geometry-centric pipelines based on emerging 3D foundation models offer a complementary capability~\cite{wang2024dust3r,zhang2024monst3r,chen2025easi3r,wang2025vggt,zhou2025page,hu2025vggt4d}. These models can recover globally coherent 4D scene structure and camera motion from monocular input, providing strong geometric priors without scene-specific training. However, such representations are not designed for high-fidelity rendering and lack the flexibility required for dynamic novel-view synthesis. As a result, current 4D systems often prioritize either geometry consistency or rendering quality, but rarely both.

This observation motivates an important question: Can dynamic Gaussian Splatting be explicitly grounded in foundation-derived geometry to achieve robust 4D reconstruction that preserves both structure and rendering quality?
In this work, we introduce Ground4D, a geometry-grounded framework that integrates robust multi-view-consistent priors from VGGT into dynamic Gaussian Splatting. Specifically, our approach is motivated by the observation that photometric optimization alone does not explicitly enforce geometric consistency across views, which can result in depth inconsistencies under viewpoint extrapolation.

We first leverage VGGT in a training-free manner to reconstruct coherent 3D geometry and camera parameters from monocular video. Since the reconstructed point cloud represents a shared scene structure, projecting these points into different camera views yields depth maps that are consistent across viewpoints. This property provides a stable geometric prior for dynamic scene modeling. The recovered 3D points are used to initialize dynamic Gaussian primitives, yielding a geometrically grounded starting representation.

\begin{figure*}[!t]
    \centering
    \includegraphics[width=\linewidth]{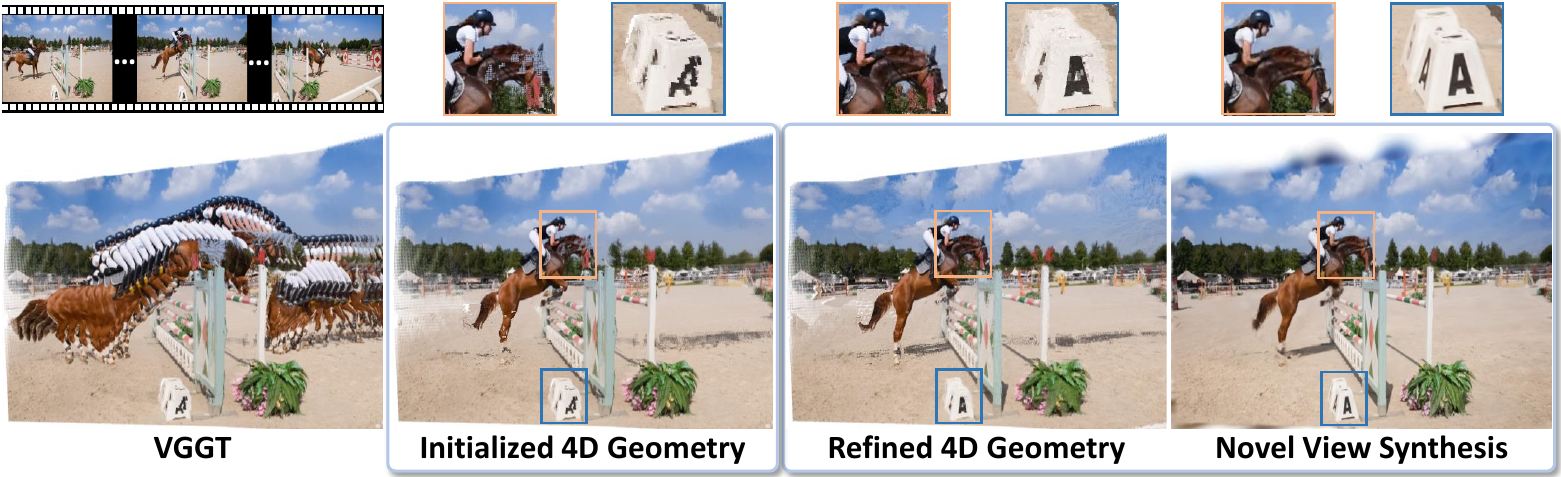}
    \caption{
    \textbf{Ground4D enables consistency-aware 4D geometry reconstruction and high-quality novel-view synthesis from a single monocular video.} While direct outputs from 3d foundation models like VGGT~\cite{wang2025vggt} provide strong spatial priors, they lack the fidelity for dynamic rendering. To bridge this gap, we first perform a training-free 4D geometry initialization to yield a geometrically grounded starting representation. We then optimize a deformable Gaussian Splatting model, maintaining multi-view geometric consistency during refinement.  Ground4D recovers high-fidelity underlying geometry and achieves photorealistic novel-view synthesis.
    \vspace{-10pt}
    }
    \label{fig:first_image}
\end{figure*}

We then optimize a deformable Gaussian Splatting model through differentiable rendering. To maintain multi-view geometric consistency during refinement, we align rendered depth with depth obtained by projecting the reconstructed 3D geometry into both observed and synthesized viewpoints along the camera trajectory. Because all depth supervision originates from the same underlying 3D structure, the optimization constrains the Gaussians to preserve coherent geometry even in regions not directly constrained by image reconstruction. After refinement, the optimized Gaussians can produce depth maps that are consistent across views and can be back-projected to obtain improved 4D scene geometry while simultaneously supporting high-quality novel-view synthesis, as demonstrated in Figure~\ref{fig:first_image}.
Furthermore, Ground4D inherently models continuous 4D dynamics, transforming discrete monocular observations into a continuous representation that enables temporally smooth novel-view synthesis and geometry reconstruction at arbitrary timesteps.
Extensive experiments on dynamic-scene benchmarks demonstrate that Ground4D improves both geometric reconstruction quality and dynamic novel-view synthesis under standard evaluation protocols.

\section{Related Work}
\label{sec:related}

\noindent\textbf{3D Foundation Models.}
Traditional Multi-View Stereo (MVS) and Structure-from-Motion (SfM)~\cite{agarwal2011building,pollefeys1999self,pollefeys2004visual,schonberger2016structure,snavely2006photo,snavely2008modeling} methods rely heavily on feature matching and bundle adjustment~\cite{agarwal2010bundle,brachmann2024scene,tang2018ba,teed2021droid,wang2024vggsfm}, which often fail in dynamic or textureless regions. Recently, 3D foundation models driven by massive data have emerged as a powerful paradigm shift. DUSt3R~\cite{wang2024dust3r} pioneered large-scale pretraining for 3D foundation models based on direct prediction by generating dense point maps from image pairs. MASt3R~\cite{leroy2024grounding} strengthened correspondence quality. However, pairwise inputs make inference costs grow quadratically with sequence length. To handle multiple views consistently, Fast3R~\cite{yang2025fast3r} and VGGT~\cite{wang2025vggt} further employ global attention for reasoning across views. Recent extensions push beyond these methods with causal streaming for long sequences~\cite{zhuo2025streaming,yuan2026infinitevggt}, permutation equivariance~\cite{wang2025pi}, and token merging acceleration for VGGT without requiring additional training~\cite{shen2025fastvggt}, delivering faster inference and stronger generalization.

\noindent\textbf{4D Geometry Reconstruction.}
While the models above excel on static scenes, extending them to dynamic 4D reconstruction remains challenging. MonST3R~\cite{zhang2024monst3r} adapts DUSt3R~\cite{wang2024dust3r} via additional training to predict point maps per timestep. CUT3R~\cite{wang2025continuous} further trains MASt3R~\cite{leroy2024grounding} into a stateful, recurrent transformer for online processing. DAS3R~\cite{xu2024das3r} trains a DPT head~\cite{ranftl2021vision} upon MonST3R for feed-forward segmentation. Easi3R~\cite{chen2025easi3r} uses the DUSt3R backbone in two passes: an initial pass analyzes cross attention to derive dynamic segments, followed by attention weight adjustment for a cleaner reconstruction. PAGE-4D~\cite{zhou2025page} extends VGGT to dynamic scenes by introducing a dynamics-aware aggregator that predicts a dynamic mask and suppresses motion cues. VGGT4D~\cite{hu2025vggt4d} mines motion cues from attention layers to disentangle dynamic objects from static backgrounds without requiring additional training. 
While these methods recover globally coherent geometry, their point-based outputs are not well suited for photorealistic rendering or dynamic novel-view synthesis. To address this limitation, we leverage VGGT’s multi-view-consistent geometry to initialize a dynamic Gaussian Splatting representation. This design connects foundation-based geometric reconstruction with differentiable volumetric rendering, allowing geometry to guide refinement within a continuous 4D representation.

\noindent\textbf{Dynamic Novel View Synthesis (NVS).}
Representing dynamic scenes for NVS has seen rapid progress, evolving from implicit Neural Radiance Fields (NeRFs)~\cite{mildenhall2021nerf,barron2021mip,liu2023robust,fu2024cbarf,liang20264dgstream} to explicit 3D Gaussian Splatting (3DGS)~\cite{kerbl20233d,yu2024mip}. Methods like 4DGS~\cite{wu20244d} and Deformable GS~\cite{yang2024deformable} introduce a deformation field to model motion. Alternatively, Dynamic 3D Gaussians~\cite{luiten2024dynamic} models motion directly by allowing Gaussians to move and rotate over time. Dynamic Gaussian Marbles~\cite{stearns2024dynamic} restricts the representation to isotropic Gaussians and employs a divide-and-conquer learning strategy guided by image-level and geometry-level priors. To handle large motions and topological changes, MoSca~\cite{lei2025mosca} proposes Motion Scaffolds, which lift 2D trajectories into a graph-based representation to drive Gaussian deformation. Shape of Motion~\cite{wang2025shape} extracts explicit, persistent 3D motion trajectories by representing scene motion with a compact set of $SE(3)$ motion bases. 
These methods primarily rely on photometric optimization and do not explicitly enforce multi-view geometric consistency, which can lead to structural inconsistencies under significant viewpoint changes. In contrast, we introduce a geometry-consistency-aware refinement stage that aligns rendered depth with projected 3D structural priors. By anchoring dynamic Gaussians to foundation-derived geometry during optimization, our approach improves geometric stability while preserving rendering fidelity.

\section{Ground4D}

\begin{figure*}[!t]
    \centering
    \includegraphics[width=\linewidth]{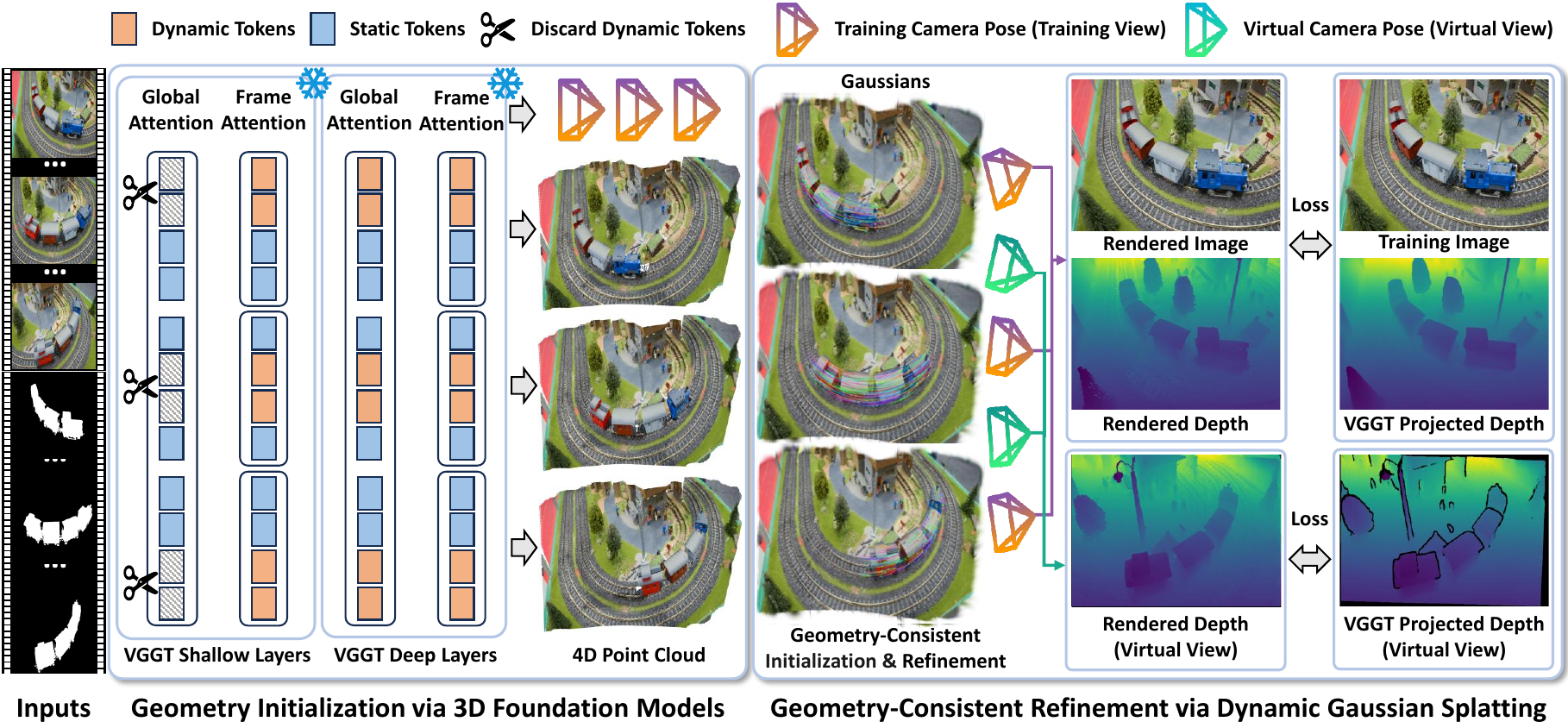}
    \caption{\textbf{Ground4D Overview.} Given a monocular video and corresponding dynamic masks extracted via SAM~\cite{kirillov2023segment}, our method proceeds in two stages. 
    (1) \textit{Geometry Initialization via 3D Foundation Models:} We leverage VGGT in a training-free manner to recover multi-view-consistent 3D geometry and camera parameters by suppressing dynamic tokens within the global attention layers. The recovered 3D points initialize dynamic Gaussian primitives, yielding a geometrically grounded starting representation. 
    (2) \textit{Geometry-Consistent Refinement via Dynamic Gaussian Splatting:} We optimize a deformable Gaussian Splatting model through differentiable rendering. To maintain multi-view geometric consistency during refinement, we align the rendered depth with the depth obtained by projecting the reconstructed 3D geometry into both observed and synthesized viewpoints along the camera trajectory. The optimized Gaussians produce view-consistent depth maps that can be back-projected to extract improved 4D scene geometry while simultaneously enabling high-quality novel-view synthesis.
    Additionally, Ground4D naturally models continuous 4D dynamics, enabling rendering and geometry reconstruction at arbitrary timesteps.
    \vspace{-15pt}
    }
    \label{fig:method}
\end{figure*}

\subsection{Overview}

\noindent\textbf{Problem Definition.}
Given a monocular video of a dynamic scene with $T$ frames $\mathcal{I} = \{I_1, I_2, \ldots, I_T\}$, our goal is twofold: (i) reconstruct globally consistent 4D scene geometry, including camera parameters $\{\mathbf{g}_t\}_{t=1}^T$, dense depth maps $\{D_t\}_{t=1}^T$, and 4D point clouds for both static and dynamic scene elements; and (ii) synthesize photorealistic images from novel viewpoints and arbitrary timesteps via high-quality novel-view rendering. 
We additionally assume access to per-frame dynamic masks $\{M_t\}_{t=1}^T$, which can be obtained via the off-the-shelf segmentation model SAM~\cite{kirillov2023segment} following~\cite{wang2025shape}.

As illustrated in Fig.~\ref{fig:method}, Ground4D proceeds in two stages:
1) \emph{Geometry Initialization via 3D Foundation models:} We leverage the Visual Geometry Grounded Transformer (VGGT)~\cite{wang2025vggt} in a training-free manner to recover multi-view-consistent 3D geometry and camera parameters from the input video. To handle dynamic content, we discard dynamic tokens within the global attention layers, preventing geometric corruption from moving objects. The recovered 3D points initialize dynamic Gaussian primitives, yielding a grounded starting representation that provides a stable geometric prior for modeling the dynamic scene; 
2) \emph{Geometry-Consistency-Aware Refinement via Dynamic Gaussian Splatting:} We optimize a deformable Gaussian Splatting model through differentiable rendering. To maintain multi-view geometric consistency during refinement, we align the rendered depth with the depth obtained by projecting the reconstructed 3D geometry into both observed and synthesized viewpoints along the camera trajectory. Because all depth supervision originates from the same 3D geometry, the optimization preserves coherent structural integrity, even in regions not directly constrained by image reconstruction. After refinement, the optimized Gaussians produce view-consistent depth maps that can be back-projected to extract improved 4D scene geometry while simultaneously enabling high-quality novel-view synthesis.

\subsection{Geometry Initialization via 3D Foundation Models}
\label{sec:geom_init}

\noindent\textbf{Visual Geometry Grounded Transformer.}
VGGT~\cite{wang2025vggt} is a large feed-forward transformer that, given a sequence of images $\{I_t\}_{t=1}^T$, predicts all key 3D attributes of the observed scene in a single forward pass:
\(
    f\left(\{I_t\}_{t=1}^T\right) = \{\mathbf{g}_t, D_t, P_t, T_t\}_{t=1}^T,
\)
where $\mathbf{g}_t \in \mathbb{R}^9$ denotes camera parameters (intrinsics and extrinsics), $D_t \in \mathbb{R}^{H \times W}$ is the depth map, $P_t \in \mathbb{R}^{3 \times H \times W}$ is the point map expressed in the coordinate frame of the first camera, and $T_t \in \mathbb{R}^{C \times H \times W}$ contains dense tracking features. For the camera parameters, $\mathbf{g}=\left[\mathbf{q}, \mathbf{t}, \mathbf{f}\right]$ which is the concatenation of the rotation quaternion $\mathbf{q} \in \mathbb{R}^4$, translation $\mathbf{t} \in \mathbb{R}^3$, and field of view $\mathbf{f} \in \mathbb{R}^2$.

Architecturally, VGGT patchifies each input image into tokens via a pretrained DINO encoder~\cite{oquab2023dinov2} and processes the combined token set through $L$ layers of Alternating Attention, which alternates between frame-wise self-attention (attending to tokens within each frame separately) and global self-attention (attending to tokens across all frames jointly). This design integrates cross-view information while maintaining per-frame normalization of activations. Camera parameters are decoded from dedicated camera tokens via a camera head, while dense outputs (depth maps, point maps) are produced through a DPT~\cite{ranftl2021vision} head.

A key property of VGGT is that its predicted point maps are viewpoint-invariant: the 3D points $P_i$ for all frames $i$ are expressed in a shared world coordinate system (the first camera frame). 
Consequently, projecting these points into any camera view yields depth maps that are inherently multi-view consistent. We exploit this property as geometric supervision in subsequent optimization.

\noindent\textbf{Training-Free 4D Geometry Reconstruction.}
While VGGT produces multi-view-consistent 3D reconstructions for static scenes, directly applying it to dynamic videos is problematic: moving objects violate the static-scene assumption underlying its global attention mechanism, leading to corrupted geometry for both dynamic and static regions.

We address this challenge with a training-free dynamic token masking strategy that modulates the shallow global attention layers to suppress the influence of dynamic content, while preserving per-frame reconstruction of moving objects through the frame-wise attention pathway. Given the per-frame dynamic masks $\{M_t\}_{t=1}^T$ obtained from SAM~\cite{kirillov2023segment}, we classify each image token as either static or dynamic based on whether the corresponding spatial region overlaps with $M_t$. 
In the global self-attention layers, where tokens from all frames attend to one another, dynamic tokens can broadcast temporally inconsistent information that corrupts the reconstruction of the static scene structure. We address this by discarding the Key and Value vectors of dynamic tokens within the first $L_s$ shallow global attention layers, while leaving the deeper layers unmodified.

Concretely, let $\mathbf{t}_t^{(l)}(k)$ denote the $k$-th token from frame $t$ in the global attention layer $l$, and let $\mathcal{S}$ and $\mathcal{D}$ denote the sets of static and dynamic tokens, respectively, determined by the masks $M_t$. For a global attention layer $l \leq L_s$, the Key and Value sets are restricted to static tokens only:
\begin{equation}
\begin{aligned}
    \tilde{\mathbf{K}}_t^{(l)}
    &= \{\mathbf{K}_t^{(l)}(k) \mid \mathbf{t}_t^{(l)}(k) \in \mathcal{S}\}, \\
    \tilde{\mathbf{V}}_t^{(l)}
    &= \{\mathbf{V}_t^{(l)}(k) \mid \mathbf{t}_t^{(l)}(k) \in \mathcal{S}\}.
\end{aligned}
\label{eq:kv_masking}
\end{equation}
while the Query vectors remain unmodified for all tokens to preserve the overall output dimensionality. 
We only mask dynamic tokens in the shallow global attention layers because fully masking dynamic tokens across the entire network would cause a severe distribution shift in the deeper feature representations.

After the forward pass, we extract the following geometric quantities for each frame $t$, where $\mathbf{y} \in \{1,\ldots,H\} \times \{1,\ldots,W\}$ denotes a pixel location:
\begin{itemize}
    \item \textbf{Camera parameters}: $\{\mathbf{g}_t\}_{t=1}^T$, provides the globally consistent camera intrinsics and world-to-camera extrinsics.
    \item \textbf{Global static point cloud}: $\mathcal{P}^\text{static} = \bigcup_{t=1}^T \{P_t(\mathbf{y}) \mid M_t[\mathbf{y}] = 0\}$.
    Because VGGT intrinsically predicts point clouds in a unified global coordinate system, we obtain the complete static background by aggregating the non-dynamic points across all frames. This yields a globally geometry-consistent background geometry without requiring additional cross-frame registration.
    \item \textbf{Per-frame dynamic point clouds}: $\mathcal{P}_t^\text{dyn} = \{P_t(\mathbf{y}) \mid M_t[\mathbf{y}] = 1\}$. These capture the geometry of dynamic objects at each specific timestep. Unlike the static point cloud, the dynamic point clouds are kept per-frame rather than aggregated, since the 3D positions of moving objects differ across timesteps.
    \item \textbf{Multi-view consistent depth maps}: For any camera viewpoint (including unobserved ones) at time $t$, depth can be obtained by projecting the reconstructed point cloud $\mathcal{P}_t=\mathcal{P}^\text{static} \cup \mathcal{P}_t^\text{dyn}$ using the estimated camera parameters: $\hat{D}_t(\mathbf{y}) = [\pi_{\mathbf{K}_t}(\mathbf{W}_t^{-1} \cdot \mathcal{P}_t)]_z$, where $\pi_{\mathbf{K}_t}$ and $\mathbf{W}_t$ denotes the perspective projection with camera intrinsics and the world-to-camera extrinsic matrix respectively, and $[\cdot]_z$ extracts the depth component.
\end{itemize}
Since all point maps share the same world coordinate system, the depth maps derived from projecting the unified 3D structure are inherently consistent across viewpoints. This multi-view consistent depth serves as a geometric anchor throughout the subsequent optimization.

\subsection{Geometry-Consistency-Aware Refinement via Dynamic Gaussian Splatting}
\label{sec:refinement}

\noindent\textbf{Deformable Gaussian Splatting.}
Dynamic Gaussian Splatting provides a differentiable 4D representation with explicit 3D parameters. When initialized from multi-view-consistent foundation geometry, the Gaussian centers inherit a coherent global structure. During optimization, gradients from rendering and geometric supervision directly update these spatial parameters, allowing the model to refine local geometry while preserving cross-view consistency.
To this end, we adopt the deformation field that transforms Gaussians across time~\cite{lei2025mosca}. 

The dynamic scene is represented as: \(\mathcal{G} = \{(\boldsymbol{\mu}_j, R_j, \mathbf{s}_j, o_j, \mathbf{c}_j;\; t_j^{\mathrm{ref}},\, \Delta\mathbf{w}_j)\}_{j=1}^N\),
where each Gaussian is parameterized by a center $\boldsymbol{\mu}_j \in \mathbb{R}^3$, a covariance matrix decomposed into a rotation $R_j$ and scale $\mathbf{s}_j$, an opacity $o_j$, and spherical harmonic (SH) coefficients $\mathbf{c}_j$ for view-dependent color and $t_j^{\mathrm{ref}}$ denotes the reference timestep at which the Gaussian is initialized and $\Delta\mathbf{w}_j$ is a per-Gaussian learnable skinning weight correction. The deformation is driven by a sparse, structured graph $(\mathcal{V}, \mathcal{E})$ whose nodes $\mathcal{V} = \{\mathbf{v}^{(m)}\}_{m=1}^{M}$ are 6-DoF trajectories encoding low rank scene motion with time-varying rigid transforms $\mathbf{Q}_t^{(m)} = [\mathbf{R}_t^{(m)}, \mathbf{t}_t^{(m)}] \in SE(3)$. Edges connect these nodes through K nearest neighbor links. To render the scene at a query time $t$, all Gaussians are deformed from their respective reference timesteps to $t$ via Dual Quaternion Blending (DQB)~\cite{kavan2007skinning} of the nearby node transformations:
\(
\mathcal{G}(t) = \{(\mathbf{T}_j(t)\boldsymbol{\mu}_j,\; \mathbf{T}_j(t) R_j,\; \mathbf{s}_j,\; o_j,\; \mathbf{c}_j)\}_{j=1}^N,
\)
where $\mathbf{T}_j(t) = \mathcal{W}(\boldsymbol{\mu}_j,\, \mathbf{w}(\boldsymbol{\mu}_j, t_j^{\mathrm{ref}}) + \Delta\mathbf{w}_j;\, t_j^{\mathrm{ref}},\, t)$ denotes the deformation field warping each Gaussian center from its reference time $t_j^{\mathrm{ref}}$ to the query time $t$, with $\mathbf{w}$ being the base radial basis function skinning weights. 
The final renderable dynamic scene is the union of the deformed foreground Gaussians $\mathcal{G}(t)$ and a static background $\mathcal{H} = \{(\boldsymbol{\mu}_j, R_j, \mathbf{s}_j, o_j, \mathbf{c}_j)\}_{j=1}^H$ represented as standard 3DGS.

\noindent\textbf{Geometry Initialization.} 
We leverage the multi-view-consistent geometry recovered from VGGT to provide a more reliable starting point. The static background Gaussians $\mathcal{H}$ are initialized by the static point cloud $\mathcal{P}^\text{static}$. Each 3D point is converted to a Gaussian primitive with position $\mu_j$ set to the point location, initial isotropic scale, and color initialized from the corresponding pixel. For the dynamic foreground, we initialize the dynamic Gaussians from the per-frame dynamic point clouds $\{\mathcal{P}_t^\text{dyn}\}_{t=1}^T$. Specifically, we use the VGGT-predicted 3D positions as the translation components of the node trajectories at each timestep.

\noindent\textbf{Geometry-Consistency-Aware Refinement.}
The VGGT-derived initialization provides a geometrically consistent starting point, but it alone does not guarantee that this consistency is preserved throughout optimization. Since photometric optimization does not explicitly enforce consistency across different views, the photometric gradients can gradually push Gaussians toward configurations that minimize image reconstruction error at training views while introducing depth inconsistencies when observed from novel viewpoints. To address this, we introduce a geometry-consistency-aware refinement that leverages the multi-view consistent depth projected from the 3D structure reconstructed by VGGT. This approach constrains the Gaussians at both observed and synthesized viewpoints throughout the entire refinement process.

For each training frame $t$, we compute the reference depth map by projecting the reconstructed geometry:
\(
    \hat{D}_t = \left[\pi_{\mathbf{K}_t}\left(\mathbf{W}_t^{-1} \cdot \mathcal{P}_t\right)\right]_z,
    \label{eq:mv_depth}
\)
where $\mathcal{P}_t = \mathcal{P}_\text{static} \cup \mathcal{P}_t^\text{dyn}$ is the combined static and dynamic point cloud at time $t$, $\mathbf{W}_t$ is the camera extrinsics, and $\pi_{\mathbf{K}_t}$ is projection with intrinsics $\mathbf{K}_t$.

Furthermore, we extend this supervision beyond the observed viewpoints. For each training viewpoint with camera parameters $\mathbf{g}_t = [\mathbf{q}_t, \mathbf{t}_t, \mathbf{f}_t]$ at time $t$, we synthesize a set of $K$ pseudo cameras $\{\tilde{\mathbf{g}}_{t,k}\}_{k=1}^K$ by perturbing the camera position and orientation: 
\(
    \tilde{\mathbf{g}}_{t,k} = \left[\, \mathbf{q}_t \otimes \Delta\mathbf{q}_k,\;\; \mathbf{t}_t + \boldsymbol{\epsilon}_k,\;\; \mathbf{f}_t \,\right],
\)
where $\Delta\mathbf{q}_k$ is a small random rotation quaternion sampled near the identity and $\boldsymbol{\epsilon}_k$ is a Gaussian perturbation to the translation. The corresponding consistent depth map is obtained via the same projection mechanism:
\(
    \hat{D}_{t,k} = \left[\pi_{\mathbf{K}_{t,k}}\left(\mathbf{W}_{t,k}^{-1} \cdot \mathcal{P}_t\right)\right]_z.
    \label{eq:synth_depth}
\)
Since both $\hat{D}_t$ and $\hat{D}_{t,k}$ originate from the same underlying 3D structure, supervising the Gaussians to match these depth maps at multiple viewpoints enforces geometric consistency across views.

Let $\tilde{D}_t$ and $\tilde{D}_{t,k}$ denote the depth rendered from the current Gaussian representation $\mathcal{G}(t) \cup \mathcal{H}$ at viewpoint $(\mathbf{W}_t, \mathbf{K}_t)$ and $(\mathbf{W}_{t,k}, \mathbf{K}_{t,k})$, respectively. We define the geometry-consistency loss as:
\begin{equation}
    \mathcal{L}_\text{gc} = \sum_{t=1}^T \left\| \tilde{D}_t - \hat{D}_t \right\| + \lambda_s \cdot \sum_{t=1}^{T} \sum_{k=1}^{K}  \left\| \tilde{D}_{t,k} - \hat{D}_{t,k} \right\|,
    \label{eq:gc_loss}
\end{equation}
where $\lambda_s$ controls the relative weight of synthesized-view supervision. The depth supervision is applied in a scale-invariant manner to account for potential scale ambiguities between the rendered and projected depth.

\noindent\textbf{Total objective.}
To further encourage neighboring nodes in the deformation field driven by $(\mathcal{V}, \mathcal{E})$ with time-varying rigid transforms $\mathbf{Q}_t^{(m)} = [\mathbf{R}_t^{(m)}, \mathbf{t}_t^{(m)}] \in SE(3)$, to move in a way that preserves local shape structure between two timesteps~\cite{lei2025mosca} by a time interval $\Delta$, we use the following three losses.
First, the as-rigid-as-possible (ARAP) loss~\cite{sorkine2007rigid,newcombe2015dynamicfusion} encourages the preservation of local distances in the neighborhood: $\mathcal{L}_\text{arap} = \sum_{t=1}^T \sum_{m=1}^M \sum_{n \in \mathcal{E}(m)} \left| \|\mathbf{t}_t^{(m)} - \mathbf{t}_t^{(n)}\| - \|\mathbf{t}_{t+\Delta}^{(m)} - \mathbf{t}_{t+\Delta}^{(n)}\| \right| + \left\| \mathbf{Q}^{-1\,(n)}_t \mathbf{t}^{(m)}_t - \mathbf{Q}^{-1\,(n)}_{t+\Delta} \mathbf{t}^{(m)}_{t+\Delta} \right\|$. 
Second, the velocity loss enforces the temporal velocity smoothness: $\mathcal{L}_\text{vel} = \sum_{t=1}^T \sum_{m=1}^M \|\mathbf{t}_t^{(m)} - \mathbf{t}_{t+1}^{(m)}\| + \|\log(\mathbf{R}_t^{(m)} \mathbf{R}_{t+1}^{-1\,(m)})\|$. 
Third, the acceleration loss regularizes the acceleration smoothness: $\mathcal{L}_\text{acc} = \sum_{t=1}^T \sum_{m=1}^M \|\mathbf{t}_t^{(m)} - 2\mathbf{t}_{t+1}^{(m)} + \mathbf{t}_{t+2}^{(m)}\| + \left| \|\log(\mathbf{R}_t^{(m)} \mathbf{R}_{t+1}^{-1\,(m)})\| - \|\log(\mathbf{R}_{t+1}^{(m)} \mathbf{R}_{t+2}^{-1\,(m)})\| \right|$.

The complete optimization objective combines the geometry-consistency loss with the photometric reconstruction loss ($\mathcal{L}_\text{rgb}$) and regularization losses:
\begin{equation}
    \mathcal{L} = \mathcal{L}_\text{rgb} + \lambda_\text{gc} \cdot \mathcal{L}_\text{gc} + \lambda_\text{r} \cdot (\mathcal{L}_\text{arap} + \mathcal{L}_\text{acc} + \mathcal{L}_\text{vel}),
    \label{eq:total_loss}
\end{equation}
where $\lambda_{\text{gc}}$ and $\lambda_{\text{r}}$ are the balancing weights for each loss term.

\noindent\textbf{Inference.}
The optimized Gaussians enable both novel-view synthesis and 4D geometry reconstruction. For rendering at query time $t$, Gaussians are deformed via the motion field and splatted. For geometry reconstruction, we back-project rendered depth maps into 3D space. The static background is aggregated across all frames, while the dynamic foreground is extracted specifically at time $t$. Their union yields a complete 4D point cloud that inherits VGGT's geometric consistency and the optimized Gaussians' photometric fidelity.

\noindent\textbf{Continuous 4D Dynamics.}
Our Ground4D naturally supports continuous 4D dynamics modeling unlike implicit dynamic GS methods~\cite{wu20244d,yang2024deformable} that parameterize motion via time-conditioned deformation fields.
Specifically, we lift the optimized discrete node transforms $\{\mathbf{Q}_t^{(m)}\}_{t=1}^{T}$ into continuous-time motion trajectories using B-spline curves:
\begin{equation}
    \mathbf{Q}_{\tau}^{(m)} = \sum_{i \in \mathcal{N}(\tau)} B_i(\tau) \mathbf{C}_i^{(m)},
    \label{eq:bspline_motion}
\end{equation}
where $B_i(\tau)$ are the B-spline basis functions, $\mathbf{C}_i^{(m)}$ are the node-specific control transforms, and $\mathcal{N}(\tau)$ denotes the local knot neighborhood at time $\tau$. 
This continuous modeling inherently enables temporally coherent novel-view synthesis and 4D geometry reconstruction at arbitrary timesteps.

\section{Experiments}

\subsection{Experimental Setup}

\noindent\textbf{Datasets.} We assess 4D geometry reconstruction and novel view synthesis on the DyCheck~\cite{gao2022monocular} dataset, the most challenging benchmark for monocular dynamic scene reconstruction. To examine robustness in real-world monocular videos, we provide qualitative results on the DAVIS~\cite{perazzi2016benchmark,pont20172017} dataset. Furthermore, camera pose estimation is evaluated on the DyCheck and TUM-Dynamics~\cite{sturm2012benchmark} datasets.

\noindent\textbf{Evaluation Metrics.} For 4D geometry reconstruction, we follow the evaluation protocol of the prior work~\cite{chen2025easi3r} and report Accuracy (mean distance from predicted to ground-truth points), Completeness (mean distance from ground-truth to predicted points), and Distance (Chamfer distance), each with mean and median variants. For novel view synthesis, we report mPSNR, mSSIM, and mLPIPS following the DyCheck protocol~\cite{gao2022monocular}. For camera pose estimation, we report Absolute Trajectory Error (ATE), Relative Translation Error (RTE), and Relative Rotation Error (RRE).

\noindent\textbf{Baselines.} For 4d geometry reconstruction and camera pose estimation, we compare against 3D foundation models including DUSt3R~\cite{wang2024dust3r}, MonST3R~\cite{zhang2024monst3r}, CUT3R~\cite{wang2025continuous}, DAS3R~\cite{xu2024das3r}, Easi3R~\cite{chen2025easi3r}, VGGT~\cite{wang2025vggt}, VGGT4D~\cite{hu2025vggt4d}, and PAGE-4D~\cite{zhou2025page}, as well as Gaussian methods MoSca~\cite{lei2025mosca} and Shape-of-Motion~\cite{wang2025shape}. For novel view synthesis, we compare against RobustDynrf~\cite{liu2023robust}, Dynamic 3D Gaussians~\cite{luiten2024dynamic}, 4DGS~\cite{wu20244d}, Gaussian Marbles~\cite{stearns2024dynamic}, Shape-of-Motion, and MoSca.

\noindent\textbf{Implementation Details.} 
Our method uses the pretrained VGGT model without any fine-tuning. We apply the training-free dynamic token masking strategy to the first $L_s = 5$ global attention layers. The total optimization steps are set to 10,000. We synthesize $K=1$ pseudo viewpoints per training view after 2,000 iterations. The loss weighting factors are set to $\lambda_{\text{s}}=0.5$ (Eq.~\ref{eq:gc_loss}), $\lambda_{\text{gc}}=0.1$, and $\lambda_{\text{r}}=0.01$ (Eq.~\ref{eq:total_loss}). All experiments are conducted on a NVIDIA A6000 GPU.

\subsection{Comparison With the State of the Art}

\begin{table}[!t]
\centering
\scriptsize
\setlength{\tabcolsep}{3pt}
\caption{\textbf{Quantitative evaluation of 4D geometry reconstruction on the DyCheck dataset~\cite{gao2022monocular}.} We compare against 3D foundation models and dynamic novel view synthesis methods. 
We report our method in two stages: our training-free geometry initialization (\textbf{Init.}) and our full geometry-consistent refinement pipeline.
}
\label{tab:geometry}
\begin{tabular}{l cc cc cc}
\toprule
 & \multicolumn{2}{c}{\textbf{Accuracy}} & \multicolumn{2}{c}{\textbf{Completeness}} & \multicolumn{2}{c}{\textbf{Distance}} \\
\cmidrule(lr){2-3} \cmidrule(lr){4-5} \cmidrule(lr){6-7}
\textbf{Method} & Mean $\downarrow$ & Median $\downarrow$ & Mean $\downarrow$ & Median $\downarrow$ & Mean $\downarrow$ & Median $\downarrow$ \\
\midrule
DUSt3R~\cite{wang2024dust3r} & 0.802 & 0.595 & 1.950 & 0.815 & 0.353 & 0.233 \\
CUT3R~\cite{wang2025continuous} & 0.458 & 0.342 & 1.633 & 0.792 & 0.326 & 0.229 \\
MonST3R~\cite{zhang2024monst3r} & 0.851 & 0.689 & 1.734 & 0.958 & 0.353 & 0.254 \\
DAS3R~\cite{xu2024das3r} & 1.772 & 1.438 & 2.503 & 1.548 & 0.475 & 0.352 \\
Easi3R$_\text{monst3r}$~\cite{chen2025easi3r} & 0.834 & 0.643 & 1.661 & 0.916 & 0.350 & 0.255 \\
Easi3R$_\text{dust3r}$~\cite{chen2025easi3r} & 0.703 & 0.589 & 1.474 & 0.586 & 0.301 & 0.186 \\
VGGT~\cite{wang2025vggt} & 0.192 & 0.115 & 1.074 & 0.300 & 0.262 & 0.143 \\
PAGE-4D~\cite{zhou2025page} & 0.259 & 0.121 & \underline{0.823} & \underline{0.236} & \underline{0.238} & 0.138 \\
VGGT4D~\cite{hu2025vggt4d} & 0.179 & 0.109 & 1.034 & 0.284 & 0.254 & 0.136 \\
MoSca~\cite{lei2025mosca} & 0.732 & 0.467 & 0.862 & 0.443 & 0.261 & 0.178 \\
Shape-of-Motion~\cite{wang2025shape} & 0.969 & 0.965 & 0.881 & 0.739 & 0.617 & 0.473 \\
\midrule
\textbf{Ground4D (Init.)} & \underline{0.136} & \underline{0.078} & 0.977 & 0.253 & 0.254 & \underline{0.126} \\
\rowcolor{myblue}
+ \textbf{Refinement} & \textbf{0.118} & \textbf{0.072} & \textbf{0.774} & \textbf{0.215} & \textbf{0.226} & \textbf{0.115} \\
\bottomrule
\end{tabular}
\end{table}

\begin{figure*}[!t]
    \centering
    \includegraphics[width=\linewidth]{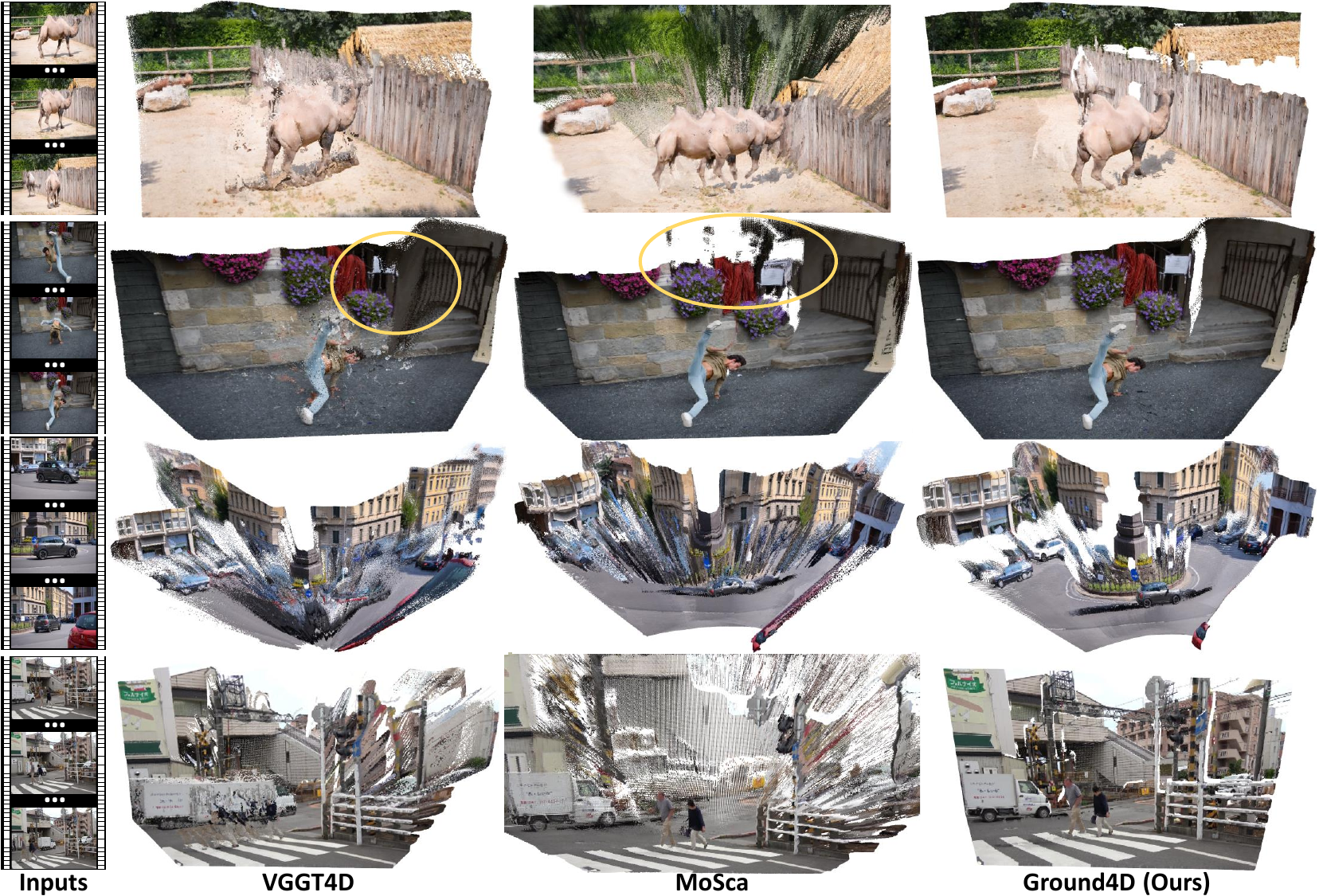}
    \caption{\textbf{Qualitative Results of the Reconstructed 4D Geometry.} Our Ground4D achieves geometry consistent reconstruction of both static scenes and moving objects.
    \vspace{-10pt}
    }
    \label{fig:qualitative_comparison1}
\end{figure*}

\noindent\textbf{4D Geometry Reconstruction.} We evaluate the geometric accuracy of reconstructed 4D scenes on the DyCheck dataset~\cite{gao2022monocular}. As shown in Tab.~\ref{tab:geometry}, we report our method's performance at two stages to highlight the contribution of each component: after our training-free geometry initialization (Init.), and after our full geometry-consistent refinement optimization.

During the initial phase, Ground4D (Init.) demonstrates strong initial performance with a mean Accuracy error of $0.136$, substantially outperforming foundation models and dynamic novel view synthesis methods by an improvement of $0.043$. This validates the effectiveness of our dynamic token masking strategy in recovering precise 4D geometry.
Notably, conventional dynamic GS methods such as MoSca and Shape-of-Motion suffer from poor geometric accuracy, confirming that relying on 2D priors yields noisy and unreliable starting geometry.
Building upon this robust initialization, our complete pipeline (+ Refinement) further optimizes the geometry to achieve the best overall results across all metrics. 
This confirms that our geometry-consistent refinement pipeline enhances the global geometric integrity of dynamic scenes.

Figure~\ref{fig:qualitative_comparison1} visualizes the reconstructed geometry across diverse dynamic scenes from the DAVIS dataset~\cite{perazzi2016benchmark,pont20172017}. 
By anchoring the optimization in VGGT's foundational 3D structure and explicitly enforcing multi-view consistency, Ground4D reconstructs dense, coherent 4D geometry with superior structural fidelity.

\begin{table}[t]
\centering
\scriptsize
\setlength{\tabcolsep}{2pt}
\caption{
\textbf{Quantitative evaluation of novel view synthesis on the DyCheck dataset~\cite{gao2022monocular}}
under three evaluation settings. Results are averaged over all applicable $7$ scenes at the standard $2\times$ resolution ($360\times480$), except Shape-of-Motion~\cite{wang2025shape}, which uses the $5$ scenes subset at $1\times$ resolution ($720\times960$). 
}
\label{tab:nvs}
\begin{tabular}{c l ccc}
\toprule
\textbf{Setting} & \textbf{Method} & \textbf{mPSNR $\uparrow$} & \textbf{mSSIM $\uparrow$} & \textbf{mLPIPS $\downarrow$} \\
\midrule
\multirow{6}{*}{With GT Pose}
 & Dynamic 3D Gaussians~\cite{luiten2024dynamic} & 7.29 & -- & 0.692 \\
 & 4DGS~\cite{wu20244d} & 13.64 & -- & 0.428 \\
 & Gaussian Marbles~\cite{stearns2024dynamic} & 16.72 & -- & 0.418 \\
 & Shape-of-Motion~\cite{wang2025shape} & 17.32 & 0.598 & 0.296 \\
 & MoSca~\cite{lei2025mosca} & \underline{19.32} & \underline{0.706} & \underline{0.264} \\
 & \cellcolor{myblue}{\textbf{Ground4D (Ours)}} & \cellcolor{myblue}{\textbf{19.43}} & \cellcolor{myblue}{\textbf{0.712}} & \cellcolor{myblue}{\textbf{0.257}} \\
\midrule
\multirow{6}{*}{Without GT Pose}
 & RobustDynRF~\cite{liu2023robust} & 17.10 & 0.534 & 0.517 \\
 & Dynamic 3D Gaussians~\cite{luiten2024dynamic} & 7.60 & -- & 0.704 \\
 & 4DGS~\cite{wu20244d} & 13.11 & -- & 0.726 \\
 & Gaussian Marbles~\cite{stearns2024dynamic} & 15.79 & -- & 0.430 \\
 & MoSca~\cite{lei2025mosca} & \underline{18.84} & \underline{0.676} & \underline{0.289} \\
 & \cellcolor{myblue}{\textbf{Ground4D (Ours)}} & \cellcolor{myblue}{\textbf{19.07}} & \cellcolor{myblue}{\textbf{0.698}} & \cellcolor{myblue}{\textbf{0.273}} \\
\midrule
\multirow{4}{*}{\makecell{Shape-of-Motion~\cite{wang2025shape}}}
 & Gaussian Marbles~\cite{stearns2024dynamic} & 16.03 & 0.543 & 0.581 \\
 & Shape-of-Motion~\cite{wang2025shape} & 16.72 & 0.630 & 0.450 \\
 & MoSca~\cite{lei2025mosca} & \underline{18.40} & \underline{0.670} & \underline{0.420} \\
 & \cellcolor{myblue}{\textbf{Ground4D (Ours)}} & \cellcolor{myblue}{\textbf{18.59}} & \cellcolor{myblue}{\textbf{0.679}} & \cellcolor{myblue}{\textbf{0.406}} \\
\bottomrule
\end{tabular}
\end{table}

\begin{figure*}[!t]
    \centering
    \includegraphics[width=\linewidth]{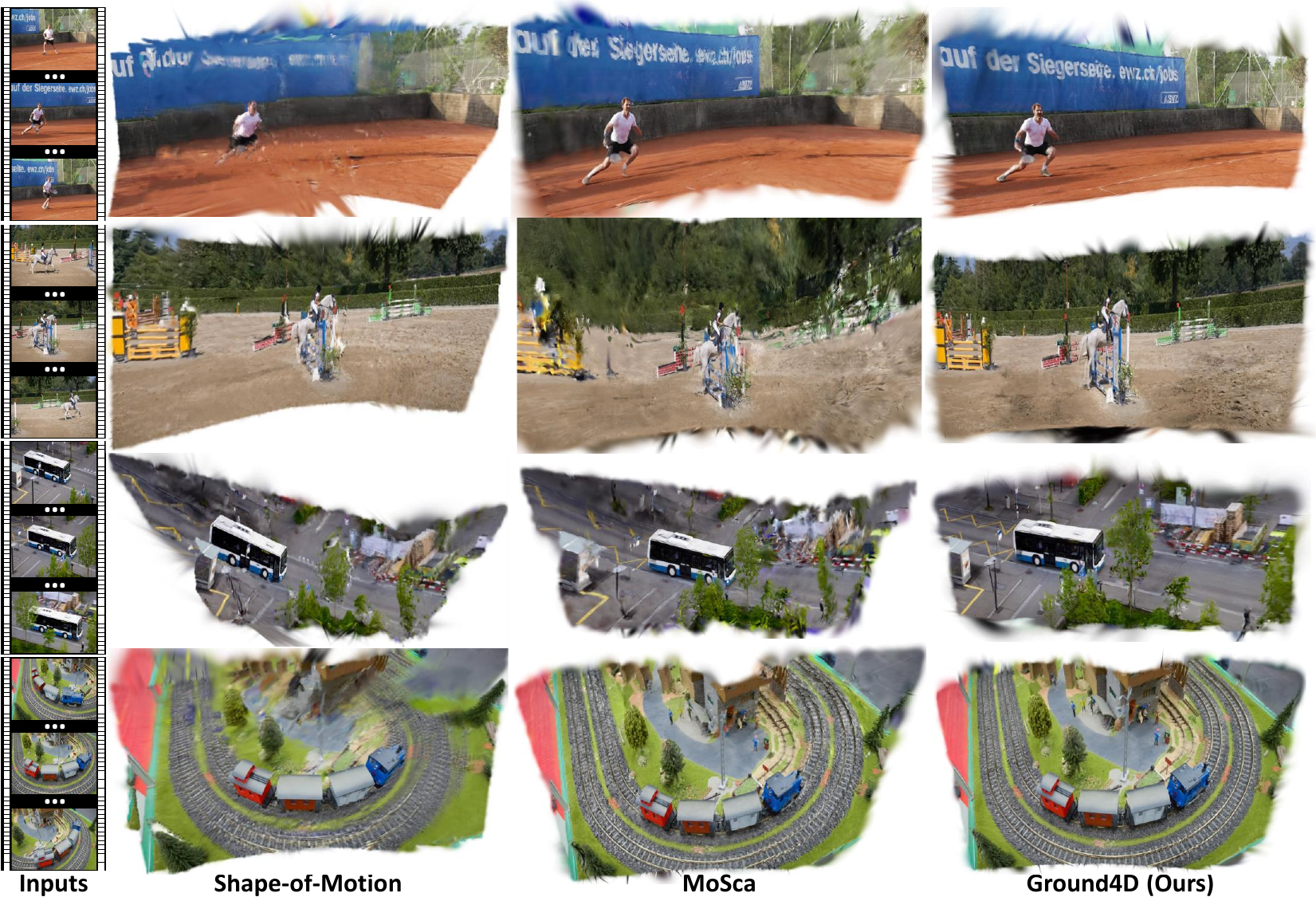}
    \caption{\textbf{Qualitative results of novel view synthesis.} 
    Ground4D effectively eliminates the blurring and artifacts, yielding sharper and more photorealistic novel views.
    }
    \label{fig:qualitative_comparison2}
\end{figure*}

\noindent\textbf{Novel View Synthesis.}
We evaluate novel view synthesis quality on the DyCheck dataset under three distinct protocols, as summarized in Table~\ref{tab:nvs}. In the standard setting with ground-truth poses, our method achieves the highest rendering fidelity, outperforming the strong baseline MoSca by mPSNR improvement of $0.11$\,dB. Crucially, in the more challenging setting without ground-truth poses, which better reflects real-world deployment where camera poses must be estimated, our method maintains the highest rendering fidelity. This highlights the advantage of our VGGT-grounded initialization, which provides reliable camera parameters and geometry even without external pose supervision. Furthermore, on the high-resolution subset used by Shape-of-Motion~\cite{wang2025shape}, our method achieves an mPSNR improvement of $0.19$\,dB, verifying that our geometrically consistent representation scales effectively to higher resolutions.

Figure~\ref{fig:qualitative_comparison2} compares novel view rendering quality across dynamic scenes from the DAVIS dataset. Methods relying on 2D priors, such as Shape-of-Motion and MoSca, struggle to maintain texture fidelity and structural coherence, often producing blurry or distorted outputs. In contrast, Ground4D synthesizes sharp, photorealistic images that perfectly preserve high-frequency details. 
By explicitly grounding dynamic Gaussians in VGGT's multi-view-consistent geometry, our approach effectively mitigates the blurring and structural artifacts in ungrounded pipelines, establishing a robust framework for dynamic novel view synthesis.

\begin{table}[t]
\centering
\scriptsize
\setlength{\tabcolsep}{5.5pt}
\renewcommand{\arraystretch}{1.1}
\caption{
\textbf{Quantitative comparisons of camera pose estimation} on the DyCheck~\cite{gao2022monocular} and TUM-dynamics~\cite{sturm2012benchmark} dataset. 
}
\label{tab:pose}
\begin{tabular}{l ccc ccc}
\toprule
 & \multicolumn{3}{c}{\textbf{DyCheck~\cite{gao2022monocular}}} & \multicolumn{3}{c}{\textbf{TUM-dynamics~\cite{sturm2012benchmark}}} \\
\cmidrule(lr){2-4} \cmidrule(lr){5-7}
\textbf{Method} & ATE $\downarrow$ & RTE $\downarrow$ & RRE $\downarrow$ & ATE $\downarrow$ & RTE $\downarrow$ & RRE $\downarrow$ \\
\midrule
DUSt3R~\cite{wang2024dust3r} & 0.035 & 0.030 & 2.323 & 0.100 & 0.087 & 2.692 \\
CUT3R~\cite{wang2025continuous} & 0.029 & 0.020 & 1.383 & 0.079 & 0.088 & 10.410 \\
MonST3R~\cite{zhang2024monst3r} & 0.033 & 0.024 & 1.501 & 0.170 & 0.155 & 6.455 \\
DAS3R~\cite{xu2024das3r} & 0.033 & 0.022 & 1.467 & 0.173 & 0.157 & 8.341 \\
Easi3R$_\text{monst3r}$~\cite{chen2025easi3r} & 0.030 & 0.021 & 1.390 & 0.168 & 0.150 & 5.925 \\
Easi3R$_\text{dust3r}$~\cite{chen2025easi3r} & 0.021 & 0.014 & 1.092 & 0.070 & 0.061 & 2.361 \\
VGGT~\cite{wang2025vggt} & 0.012 & 0.016 & 0.705 & 0.013 & 0.019 & 0.686 \\
PAGE-4D~\cite{zhou2025page} & 0.015 & 0.019 & 1.079 & 0.027 & 0.028 & 0.742 \\
VGGT4D~\cite{hu2025vggt4d} & \underline{0.011} & \underline{0.014} & \underline{0.624} & \underline{0.013} & \underline{0.019} & \underline{0.682} \\
MoSca~\cite{lei2025mosca} & 0.024 & 0.021 & 0.877 & 0.034 & 0.035 & 5.385 \\
Shape-of-Motion~\cite{wang2025shape} & 0.159 & 0.065 & 5.821 & 0.157 & 0.239 & 6.862 \\
\rowcolor{myblue}
\textbf{Ground4D (Ours)} & \textbf{0.010} & \textbf{0.013} & \textbf{0.610} & \textbf{0.012} & \textbf{0.018} & \textbf{0.669} \\
\bottomrule
\end{tabular}
\end{table}

\noindent\textbf{Camera Pose Estimation.} Table~\ref{tab:pose} reports camera pose estimation accuracy on both the DyCheck and TUM-dynamics datasets. Our method surpasses the base VGGT and its dynamic extension VGGT4D, achieving the lowest Absolute Trajectory Error (ATE) of $0.010$ on DyCheck. This improvement validates that our dynamic attention suppression strategy effectively filters out motion-induced noise, enabling the global attention mechanism to focus on static background cues more precisely.
In contrast, methods relying on 2D priors often struggle in dynamic environments. These results confirm that grounding the system in a robust, multi-view-consistent foundation model provides a clear advantage for estimating camera ego-motion in complex dynamic scenes.

\subsection{Evaluation and Analysis}

\begin{figure*}[!t]
    \centering
    \includegraphics[width=\linewidth]{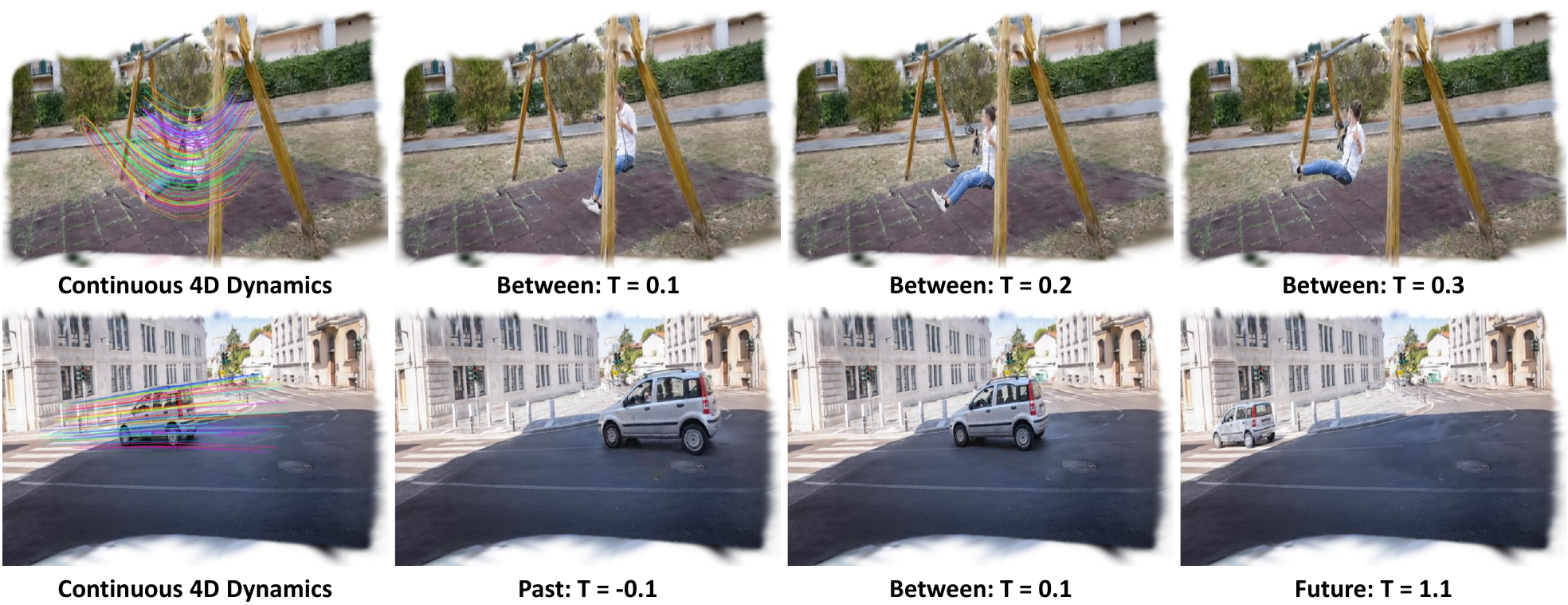}
    \caption{\textbf{Visualization of continuous-time novel-view synthesis.} 
    The temporal range of each input video is normalized to $[0,1]$. Ground4D queries the learned continuous 4D dynamics at arbitrary timestamps to render dynamic scenes from novel viewpoints. The results show smooth foreground motion and coherent scene geometry, demonstrating that Ground4D lifts discrete frame-wise 4D reconstruction into continuous-time dynamic rendering.
    }
    \label{fig:continuous_time}
\end{figure*}

\noindent\textbf{Continuous-Time Novel-View Synthesis.} We further evaluate the continuous-time rendering capability of Ground4D through qualitative visualization. We normalize the temporal range of the input video to $[0,1]$ and query the learned continuous 4D dynamics at arbitrary timestamps $\tau$. As shown in Figure~\ref{fig:continuous_time}, Ground4D synthesize novel-view results not only at intermediate timestamps within the observed temporal range, but also at extrapolated past and future timestamps (e.g., $\tau < 0$ and $\tau > 1$). The visualized motion trajectories and rendered frames show that the continuous node-based dynamics produce smooth foreground evolution while maintaining stable scene geometry and appearance. These results indicate that Ground4D naturally lifts the optimized discrete 4D representation into a continuous-time dynamic model, enabling novel-view synthesis across both interpolated and extrapolated temporal queries.

\begin{table}[t]
\centering
\scriptsize
\setlength{\tabcolsep}{4pt}
\renewcommand{\arraystretch}{1.1}
\caption{\textbf{Ablation study on training-free dynamic token masking layers.} We evaluate the impact of masking dynamic tokens across different global attention layers on camera pose estimation and geometry (Chamfer Distance) on DyCheck.}
\label{tab:ablation_masking}
\begin{tabular}{l ccc cc}
\toprule
\textbf{Training-Free Token Masking} & \multicolumn{3}{c}{\textbf{Camera Pose Estimation}} & \multicolumn{2}{c}{\textbf{Geometry Distance}} \\
\cmidrule(lr){1-1} \cmidrule(lr){2-4} \cmidrule(lr){5-6}
Masking Layers & ATE $\downarrow$ & RTE $\downarrow$ & RRE $\downarrow$ & Mean $\downarrow$ & Median $\downarrow$ \\
\midrule
No Masking & 0.012 & 0.016 & 0.705 & 0.262 & 0.143 \\
Layers 1--4 & 0.011 & 0.014 & 0.619 & 0.256 & 0.128 \\
\rowcolor{myblue} %
\textbf{Layers 1--5} & \textbf{0.010} & \textbf{0.013} & \textbf{0.610} & \textbf{0.254} & \textbf{0.126} \\
Layers 1--6 & \textbf{0.010} & 0.014 & 0.614 & 0.255 & \textbf{0.126} \\
Full Masking & 0.040 & 0.031 & 2.719 & 0.414 & 0.285 \\
\bottomrule
\end{tabular}
\end{table}

\noindent\textbf{Effectiveness of Dynamic Token Masking.} 
Table~\ref{tab:ablation_masking} ablates the masked global attention layers to validate our initialization strategy. No Masking (vanilla VGGT) exhibits degraded geometry, likely due to dynamic feature interference across views. Full Masking (masking Layers 1--24) significantly reduces performance, suggesting that excessive suppression impairs the model’s ability to aggregate useful global information. Restricting masking to shallow layers strikes a better balance: it filters early motion-induced noise while retaining higher-level contextual reasoning. Masking Layers 1--5 achieves the best initialization for both camera poses and geometry. 

\begin{table}[t]
\centering
\scriptsize
\setlength{\tabcolsep}{1pt}
\renewcommand{\arraystretch}{1.1}
\caption{\textbf{Ablation study on geometry-consistent refinement.} We evaluate the impact of enforcing multi-view depth consistency on observed (training view) and synthesized (virtual view) viewpoints on DyCheck.}
\label{tab:ablation_geometry}
\begin{tabular}{cc cccc cc}
\toprule
\multicolumn{2}{c}{\textbf{Geometry-Consistency Refinement}} & \multicolumn{3}{c}{\textbf{Novel View Synthesis}} & \multicolumn{2}{c}{\textbf{Geometry Distance}} \\
\cmidrule(lr){1-2} \cmidrule(lr){3-5} \cmidrule(lr){6-7}
Training View & Virtual View & mPSNR$\uparrow$ & mSSIM$\uparrow$ & mLPIPS$\downarrow$ & Mean$\downarrow$ & Median$\downarrow$ \\
\midrule
             &              & 18.72 & 0.688 & 0.282 & 0.261 & 0.142 \\
\checkmark   &              & 18.91 & 0.692 & 0.276 & 0.237 & 0.121 \\
\rowcolor{myblue}
\checkmark   & \checkmark   & \textbf{19.07} & \textbf{0.698} & \textbf{0.273} & \textbf{0.226} & \textbf{0.115} \\
\bottomrule
\end{tabular}
\end{table}

\noindent\textbf{Effectiveness of Geometry-Consistent Refinement.}
Table~\ref{tab:ablation_geometry} ablates the effect of depth supervision for geometry-consistent refinement. Solely on photometric optimization struggles with rendering and geometric fidelity. Enforcing geometric consistency on observed training views improves structural anchoring. Furthermore, extending these constraints to unobserved, virtual viewpoints yields the most significant gains. By regularizing synthesized views, full model achieves the best novel view synthesis and geometry.

\begin{table}[t]
\centering
\scriptsize
\setlength{\tabcolsep}{6pt}
\renewcommand{\arraystretch}{1.1}
\caption{\textbf{Ablation study on mask source.} We evaluate the impact of using dynamic masks from different sources on geometry reconstruction accuracy and completeness.}
\label{tab:ablation_mask_source}
\begin{tabular}{cccc}
\toprule
\textbf{Method} & \textbf{Mask Source} & \textbf{Mean Accuracy} $\downarrow$ & \textbf{Mean Completeness} $\downarrow$ \\
\midrule
VGGT4D   & VGGT4D & 0.179 & 1.034 \\
VGGT4D   & SAM    & 0.177 & 1.023 \\
\midrule
Ground4D & VGGT4D & 0.137 & 0.981 \\
Ground4D & SAM & \textbf{0.136} & \textbf{0.977} \\
\bottomrule
\end{tabular}
\end{table}

\noindent\textbf{Effectiveness of Mask Source.}
Table~\ref{tab:ablation_mask_source} studies the influence of different dynamic mask sources on geometry reconstruction.  
When compared under identical mask sources, Ground4D consistently outperforms VGGT4D. Our method achieves notably lower errors in both mean accuracy and mean completeness, highlighting the inherent robustness and superiority of our geometry modeling strategy.
This demonstrates that the improvement mainly comes from our geometry modeling strategy rather than a specific mask source.

\begin{table}[t]
\centering
\scriptsize
\setlength{\tabcolsep}{5pt}
\renewcommand{\arraystretch}{1.1}
\caption{\textbf{Ablation study on virtual view sampling.} We evaluate the impact of the number of sampled virtual views $K$ and the camera perturbation range for geometry-consistent refinement.}
\label{tab:ablation_virtual_view}
\begin{tabular}{ccccc}
\toprule
\textbf{$K$} & \textbf{Rot. Max Angle} & \textbf{Trans. Max Ratio} & \textbf{mPSNR} $\uparrow$ & \textbf{Mean Distance} $\downarrow$ \\
\midrule
0 & $0^\circ$  & 0   & 18.91 & 0.237 \\
\midrule
1 & $5^\circ$  & 0.1 & \textbf{19.07} & 0.226 \\
1 & $10^\circ$ & 0.2 & 19.02 & 0.228 \\
2 & $5^\circ$  & 0.1 & 19.04 & \textbf{0.225} \\
\bottomrule
\end{tabular}
\end{table}

\noindent\textbf{Effectiveness of Virtual View Sampling.}
Table~\ref{tab:ablation_virtual_view} ablates the virtual view number ($K$) and rotation/translation (Rot./Trans.) perturbation magnitudes. 
Compared with the setting without virtual views, all virtual-view configurations improve both rendering quality and geometry accuracy, confirming the effectiveness of synthesized-view geometric constraints. 
The performance varies only slightly across different perturbation ranges and $K$ values, showing that the refinement is robust to these hyperparameters. 

\section{Conclusion}

{
In this work, we presented Ground4D, a geometry-grounded framework for dynamic 4D reconstruction from monocular video. Our approach combines foundation-based geometry initialization with consistency-aware refinement through dynamic Gaussian optimization, enabling coherent 3D reconstruction alongside high-quality novel-view synthesis within a single modeling pipeline. Rather than using foundation-derived geometry solely as an initialization, Ground4D maintains multi-view-consistent structure as a persistent constraint during refinement. This design stabilizes differentiable optimization and preserves cross-view structural coherence while retaining the expressive flexibility of dynamic Gaussian representations.
Extensive experiments demonstrate that grounding dynamic scene optimization in reliable geometric priors improves both reconstruction accuracy and rendering quality. This suggests that integrating foundation-level geometry with learnable 4D representations is a promising direction for robust monocular dynamic scene modeling.
}

\noindent\textbf{Potential limitations and future directions.} 
{
Ground4D relies on a 3D foundation model for geometry initialization, which may become computationally demanding for very long video sequences. Improving efficiency through key-frame selection~\cite{sheng2019unsupervised,shen2025fastvggt} or incremental geometry propagation~\cite{chen2025ttt3r,yuan2026infinitevggt} is a promising direction. In addition, our current implementation uses SAM for dynamic–static separation; future work could instead leverage intrinsic representations within the foundation model~\cite{chen2025easi3r,hu2025vggt4d} to achieve tighter integration. 
}

\bibliographystyle{IEEEtran}
\bibliography{main}

\end{document}